\documentclass[11pt,a4paper]{article}
\usepackage[hyperref]{emnlp2020}
\usepackage{times}
\usepackage{latexsym}
\usepackage{cleveref}
\usepackage{graphicx}

\usepackage{microtype}

\aclfinalcopy 

\title{Extractive Summarizer for Scholarly Articles}

  \author{ Athar Sefid \\
Penn State University \\
\texttt{azs5955@psu.edu} \\
 \\\And
 Clyde Lee Giles \\
 Penn State University \\
\texttt{clg20@psu.edu} \\
\\\And
Prasenjut Mirta \\
 Penn State University \\
\texttt{pum10@psu.edu }\\}

\date{}

\begin{document}
\maketitle
\begin{abstract}
  We introduce an extractive method that will summarize long scientific papers. Our model uses presentation slides provided by the authors of the papers as the gold summary standard to label the sentences. The sentences are ranked based on their novelty and their importance as estimated by deep neural networks. Our window-based extractive labeling of sentences results in the improvement of at least 4 ROUGE1-Recall points. 
\end{abstract}
\section{Introduction}
The growing literature is one substantial challenge scholars face. 
As such there has been research to improve search, retrieval, and summarization of scholarly papers and to apply natural language understanding to scholarly text.

Much previous work on summarization focuses on generating headlines for news articles or generating an abstract. This is different from a summary of a scholarly document which should cover all of the significant aspects and facts of a paper.   

Our contribution is two-fold: 
\begin{itemize}
    \item  Introduce a window-based labeling approach and compare the performance of state of the art extractive summarization methods on scientific articles. 
    \item Use presentation slides generated by the authors of papers as a gold summary standard to train our model. 
\end{itemize}

\section{Related Work}
\subsection{Summarization}
Abstractive \cite{see-etal-2017-get, paulus2017deep} and extractive methods \cite{bae2019summary, zhong2020extractive} are the two main summarization techniques. An abstractive method  paraphrases the input text and its vocabulary is not limited to the source document. However, it does not always result in grammatically/factually correct sentences. Producing a correct and robust summary which is consistent with the original article is what is desired for the scientific document summarization task. 

Before deep neural networks, graph-based methods were one of the main text summarization techniques. Methods such as TextRank \cite{mihalcea2004textrank} model the sentences of the input text as nodes of the graph and the edges are labeled based on the similarity of the sentences. TextRank then applies the Google PageRank \cite{page1999pagerank} algorithm to rank the sentences. 
Many of the traditional machine learning methods apply integer linear programming to optimally select the ranked sentences by considering their length and salience \cite{galanis2012extractive}. 

SummaRuNNer \cite{nallapati2017summarunner} is a neural extractive summarizer that considers the summarization task as a sequence labeling problem that labels the sentences with a sequence of zeros and ones. The model sequentially ranks the sentences based on their position in the document, their importance with respect to the full document embedding, and novelty compared to the current state of the summary. The structure of our model is based on SummaRuNNer. However, it tries to improve the extractive labeling approach they use for summarization of long scientific papers. Our model handles the problem of the sparsity of the positively labeled sentences by manually increasing the weight of positive sentences in the cross-entropy loss. 

Recent work by Al-Sabahi et. al \cite{al2018hierarchical} leverages attention weights on both word level and sentence level to represent documents. They claim that the hierarchical attentional document representation is able to identify important sections of the document that needs to be present in the summary.    




\subsection{Automatic Presentation Generation}
There is also work \cite{hu2014ppsgen,sefid2019automatic} that  automatically generates presentation slides from scientific papers. The PPSGen \cite{hu2014ppsgen} project crawled a collection of 1200 paper/slides pairs from the websites of authors which we will use for training our summarizers.

\section{Method}

Given a document $D= s_1, s_2, ... s_n$ with $n$ sentences and extractive labels $Y = y_1, y_2, ... y_n$ where $y_i \in [0,1]$, the system predicts $p(y_i=1)$, necessity of sentence $i$ for the document summary.

\subsection{Labeling}

Gold standard summaries are abstractive summaries that describe the content of the input document. The conversion of abstractive summaries to extractive labels for the sentences is challenging and adds another layer of estimation to our approach. 

SummaRuNNer sequentially labels the sentences. If adding the sentence to the summary improves the ROUGE score, the sentence is labeled with 1, otherwise it is labeled with 0. This method of labeling is suitable for news articles such as CNN/DailyMail \cite{nallapati2016abstractive} where the first couple of sentences in articles usually cover the main content, and therefore beating the summary built by just the first three sentences (lead3) is considered to be challenging. However, in scholarly articles with multiple sections, each section should have its own portion in the summary and the labeling process should be adapted to have more distributed positive labels across sections of the paper.

In order to mitigate the labeling issue, we designed a window-based labeling approach. A window contains $w$ consecutive sentences and the sentence with the best ROUGE score in the window is labeled with 1.
Then the window slides to cover the next $w$ sentences. This way, the positive labels are scattered across the paper and its sections. A window size of $10$ is used since we found empirically it gives the best total ROUGE score (\Cref{window}). 

\begin{table}
  \caption{ROUGE Scores for different window sizes. 10 sentences in windows results in the best average ROUGE recall score on the validation set.}
  \label{window}
  \centering
  
  \begin{tabular}{c|c}
  \hline
    Window Size &  ROUGE-1 recall\\
    
    \hline
    3 & 41.22\\
    5 & 42.12\\
    7 & 43.24\\
    10 & 44.97\\
    15 & 43.85\\
  \end{tabular}
\end{table}
\section{Ranking}

The ranking of sentences mainly depends on their salience, novelty, and content. These factors are estimated based on document embedding and the embedding of the summary. This section elaborates on two different ways to represent a document and explains how the summary is embedded based on previous decisions of the model.

\subsection{Simple Document Embedding}
\newcommand{\cev}[1]{\reflectbox{\ensuremath{\vec{\reflectbox{\ensuremath{#1}}}}}}
A simple document embedding can be the average of the sentence embedding produced by bi-directional LSTMs (biLSTMs) \cite{hochreiter1997long}. A biLSTM is a Long Short Term Memory network designed to remember long term dependencies both from the beginning and end of the sentence. 

A sentence embedding is:
\begin{equation}
    \small
    E_{s_i} = [\vec{h_i},\cev{h_i}]
\end{equation}
Where $\vec{h_i}$ and $\cev{h_i}$ are the forward and backward hidden states of the biLSTM for sentence $s_i$. 

The document embedding would be the average of the sentence emebeddings: 
\begin{equation}
    \small
    E_D = ReLU( W* \frac{1}{n} \sum_{i=1}^n E_{s_i} + b)
\end{equation}
where $E_D$ is the embedding for document $D$ with $n$ sentences, $ReLU$ is the activation function, and W is the parameter to learn.  

\subsection{Hierarchical Self Attention Document Embedding}

We discuss how to represent a document by applying attention layers on both word and sentence levels \cite{al2018hierarchical, yang2016hierarchical}.

\subsubsection{Sentence embedding}

Sentence embeddings are formed by running recurrent neural networks (here a LSTM) in both forward and backward directions and then adding an attention layer on top of hidden vectors of the recurrent layer.

Consider the sentence $i$ with $m$ words:

\begin{equation}
    \small
    s_i = [w_1, w_2, ..., w_j ... , w_m]
\end{equation}
After running LSTM on $s_i$, the hidden state $h_j$ at state $j$ would be:
\begin{equation}
    \small
    h_j = [\vec{h_j},\cev{h_j}]
\end{equation}
which is the concatenation of forward ($\vec{h_j}$) and backward ($\cev{h_j}$) hidden states of the LSTM cell at step j. 
The $h_{s_i}$ is defined as a combination of all m hidden states:
\begin{equation}
    \small
    h_{s_i} = [h_1, h_2, ..., h_m]
\end{equation}
where $h_{s_i} \in R^{m \times 2d}$, $m$ is number of words in the sentence, and $d$ is the embedding dimension for each word.
The attention weights are:
\begin{equation}
    \small
    a_{word} = sotfmax(W_{attn} * h_{s_i}^T)
\end{equation}
where $ W_{attn} \in R^{k\times2d}$ is a learnable parameter. Then $a_{word} \in R^ {k \times m}$ and the embedding for sentence $s_i$ ($E_{s_i}$) would be:
\begin{equation}
    \small
    E_{s_i} = a_{word} * h_{s_i}
\end{equation}
where $E_{s_i} \in R^{k \times 2d}$.
\subsubsection{Document Embedding}
 We employ a similar approach used for sentence embedding in order to obtain the document embeddings. The document embedding has a biLSTM layer applied to the sentence embeddings ($E_{s_i}$) built in the previous step and then has another attention layer to make the document vectors.
 
\begin{figure}[htbp]
\includegraphics[width=0.50\textwidth, height=3.4cm]{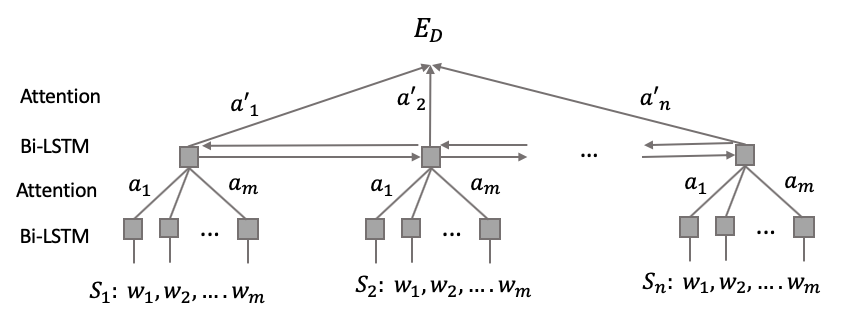}
\caption{Hierarchical self attention model applies the bi-LSTM and attention layers at the word and sentence levels.}
\label{hat}
\end{figure}

 A document $D$ is defined as:
\begin{equation}
    \small
    D = {s_1, s_2, ..., s_i, ... s_n}
\end{equation}
After the biLSTM is used on the sentence embeddings ($E_{s_i}$) from the previous step, the hidden state $h^{'}s_i$ at state $i$ would be:
\begin{equation}
    \small
    h^{'}s_i = [\vec{h^{'}s_i},\cev{h^{'}s_i} ]
\end{equation}
Again $h^{'}s_i$ is the concatenation of forward ($\vec{h^{'}s_i}$) and backward ($\cev{h^{'}s_i}$) hidden states. All of the n hidden states are combined to make $h_D$:
\begin{equation}
    \small
    h_D = [h^{'}s_1, h^{'}s_2, ..., h^{'}s_n ]
\end{equation}
The attention weights $a_{sent} \in R^{k\times n} $ learn which sentences are more important in the document representation and are built as follows:
\begin{equation}
    \small
    a_{sent} = sotfmax(W'_{attn} * h_{D}^T)
\end{equation}
where $W'_{attn} \in R^{k \times n}$. The document vector is the weighted sum of its sentence embeddings:
\begin{equation}
    \small
    E_D = a_{sent} * h_{D}
\end{equation}
This hierarchical embedding is shown in \Cref{hat}.
\begin{table*}
  \caption{ROUGE Scores for different models. SummaRuNNer with the window-based labeling improves the ROUGE score by at least 4 points}
  \label{ROUGE_score}
  \centering
  
  \begin{tabular}{c c c c c}
  \hline
    score & ROUGE-1 recall &  ROUGE-2 recall & ROUGE-L recall & training time \\
    \hline
    lead 20\% & 34.37 & 9.2 & 18.3 & -\\
    TextRank & 31.87 & 8.28 & 17.75& - \\
    SummaraRuNNer & 37.60 & 9.4 & 18.6 & 18 hours\\
    Attention + SummaruRuNNer & 37.50 & 9.44 & 17.89 & 38 hours\\
    windowed SummaRuNNer & \textbf{42.2} & \textbf{11.32} & \textbf{21.15} & 18 hours \\
  \end{tabular}
\end{table*}
\subsection{Sentence Ranking}

Our ranking is borrowed from SummRuNNer. The ranking of a sentence depends on its position in the paper. We use positional embedding to represent the $position$ of the sentence in the document: 

\begin{equation}
    \small
    pos = position * W_{pos}
\end{equation}
\begin{equation}
    \small
    salience = E_D * W_{salience}* E_{s_i}
\end{equation}
\begin{equation}
    \small
    novelty = Summary_i * W_{novelty}* E_{s_i}
\end{equation}
\begin{equation}
    \small
    p(y_i =1) = \sigma(pos+ E_{s_i} + novely + salience) 
\end{equation}
where $position$ is the relative position of the sentence in the document and $pos$ is its positional embedding. $salience$ is a function estimator for the importance of the sentence compared to the whole document. $novelty$ represents the novelty of the sentence with respect to the current summery built until $i$th sentence. 
All of $W_{pos}$, $W_{salience}$ $W_{novelty}$ are the parameters to be learned by the model and $\sigma$ is the sigmoid activation function.
\subsection{Summary Embedding}
\begin{figure}[htbp]
\includegraphics[width=0.48\textwidth, height=4cm]{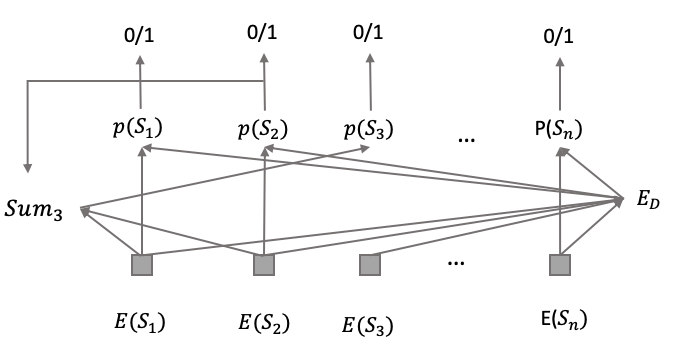}
\caption{Score prediction for sentence 3 depends on document embedding ($E_D$), sentence embedding, the embedding of the summary built until step 3 ($Sum_3$), and position of the sentence which is 3. The summary is the weighted sum of the embeddings of the first and second sentences. }
\label{sumarunner}
\end{figure}

The summary embedding is the weighted sum of the sentences added to summary:
\begin{equation}
    \small
    summary_i = \sum_{j=0}^{i-1} p(y_i=1) * E_{s_i}
\end{equation}
The higher chance of adding the sentence to the summary gives it a bigger portion in the summary embedding. 
\Cref{sumarunner} shows the architecture for predicting the score for the third sentence.
\subsection{Loss Function}
The loss function is the cross-entropy loss. It minimizes the negative log likelihood of labels and is defined as:
\begin{equation}
    \small - \sum_{i=0}^{n} w_1 y_i * log (p(y_i=1)) + w_2 (1-y_i) * log (1-p(y_i=1))
\end{equation} 
With window-based labeling, the positive labels are sparse. To deal with the unbalanced data we added weights  $w_1$ and $w_2$ to the positive and negative labels in the cross-entropy loss. Setting $w_1$ to be much larger than $w_2$ helps the model to predict and learn the important features. The setting of $w_1=-85$ and $w_2=-2$ results in the best model with the highest ROUGE score.

\section{Experiments and Results}

The dataset~\cite{hu2014ppsgen} contains 1200 academic papers in the field of computer science and the presentation slides made by the authors of the articles. The papers are in PDF format and are converted to a text file in TEI (Text Encoding Initiative) format by GROBID \cite{GROBID}. The slides are either in PDF or PPT. The LibreOffice pdftotext tool is used to convert them to text. A portion of 1000, 100, and 100 documents are used for training, validation, and testing, respectively. 

Stanford CoreNLP \cite{manning-EtAl:2014:P14-5} is used to tokenize and lemmatize sentences to the constituent words. 
GloVe 1.2 \cite{pennington2014glove} 50-dimensional vectors are used to initialize the word embeddings.
With the AdaDelta optimizer and a learning rate of 0.1, we trained for 50 epochs. The sentences are truncated or padded to have 50 tokens. In the same way, documents are set to have a fixed size of 500 sentences. There are only 12 documents in our dataset with more than 500 sentences. 

The standard ROUGE score \cite{lin2004rouge} is used to evaluate the summaries. ROUGE-N is an $n$-gram overlap between a candidate summary and a reference summary and the Longest Common Sub-sequence (LCS) of a summary and a document is represented in ROUGE-L score. The ROUGE scores for summaries calculated by Py-ROUGE package \footnote{https://pypi.org/project/py-rouge/} are tabulated in \Cref{ROUGE_score}.  

Based on our results, the hierarchical attention model with more parameters to learn and longer training time does not improve the recall or the coverage of the summaries. On the other hand, the window-based labeling technique helps the model identify important sentences across all sections of the paper and outperforms the other models by at least 4 points for ROUGE-1. 

\section{Conclusion}
We use presentation slides created by authors as a foundation for document summarization. A large source of crawled slides is used as training data.
State of the art deep learning extractive summarization methods are used to summarize papers. Our results show that distributing the positive labels across all sections of the paper in contrast with summarization methods for news articles considerably improves performance. 

Future work would be to design a system to automatically crawl for author papers and presentations. It would be interesting to use reinforcement learning to integrate cross-entropy loss with rewards from a policy gradient to optimize the ROUGE evaluation metric \cite{narayan2018ranking}. 
Our full code will be available after acceptance.

\bibliographystyle{acl_natbib}
\bibliography{anthology,emnlp2020}
\end{document}